\pdfoutput=1
\documentclass[11pt]{article}

\usepackage[]{acl}

\usepackage{times}
\usepackage{latexsym}
\usepackage{graphicx}

\usepackage[T1]{fontenc}
\usepackage[utf8]{inputenc}

\usepackage{microtype}

\newcommand{\nop}[1]{}

\title{An Effective, Performant Named Entity Recognition System for \\ Noisy Business Telephone Conversation Transcripts}

\author{Xue-Yong Fu, \ Cheng Chen, \ Md Tahmid Rahman Laskar, \\ {\bf Shashi Bhushan TN, \ Simon Corston-Oliver}\\
        Dialpad Canada Inc.\\ Vancouver, BC, Canada\\
        \texttt{\{xue-yong,cchen,tahmid.rahman\}@dialpad.com}\\\texttt{\{sbhushan,scorston-oliver\}@dialpad.com}}

\begin{document}

\maketitle

%\vspace{-2mm}

\begin{abstract}
%\vspace{-1.5mm}

We present a simple yet effective method to train a named entity recognition (NER) model that operates on business telephone conversation transcripts that contain noise due to the nature of spoken conversation and artifacts of automatic speech recognition. We first fine-tune LUKE, a state-of-the-art Named Entity Recognition (NER) model, on a limited amount of transcripts, then use it as the teacher model to teach a smaller DistilBERT-based student model using a large amount of weakly labeled data and a small amount of human-annotated data. The model achieves high accuracy while also satisfying the practical constraints for inclusion in a commercial telephony product: realtime performance when deployed on cost-effective CPUs rather than GPUs.

%Telephone transcription data can be very noisy due to speech recognition errors, disfluencies, repetition, and many other factors. It makes the property of such data fundamentally different from the data that most pre-trained models are trained on. Thus, to fine-tune pre-trained models on such data, it requires a large amount of in-domain noisy data to be annotated for a downstream task. In this paper, we present a simple yet effective method to largely reduce the need of data annotation for noisy datasets. More specifically, we first fine-tune LUKE, a state-of-the-art Named Entity Recognition (NER) model, on a limited amount of noisy telephone conversations to use it as the teacher model to teach a smaller DistilBERT-based (of-the-shelf DistilBERT) student model via large amount of weakly labeled and small amount of labeled data. The proposed approach helps the student model to achieve 213x inference speed boost while reserving 99.14\% F1 score of its teacher. We deploy our distilled model for real time NER on our communication-as-a-service (CaaS) system. 

\end{abstract}
%\vspace{-3mm}
\section{Introduction}

We describe a named entity recognition (NER) system that identifies entities mentioned in English business telephone conversations. The input to the NER system is transcripts produced by an automatic speech recognition (ASR) system. These transcripts are inherently noisy due to the nature of spoken communication and due to the limitations of the ASR system. The transcripts contain dysfluencies, false starts, filled pauses, they lack punctuation information and have incomplete information about case.

Because there was no pre-existing annotated data set publicly available that matched the characteristics of the ASR transcripts in the domain of business telephone conversations \cite{li2020survey}, the NER model is required to be trained on a large dataset containing telephone conversations to effectively detect named entities in such noisy data. % % is required to be trained using a small amount of human annotated in-domain data while leveraging pre-existing resources. % we trained a model using pre-existing resources that we fine-tuned using a small amount of human-annotated in-domain data.
Moreover, the NER model needs to provide realtime functionality in a commercial communication-as-a-service (CaaS) product such as displaying information related to the named entities to a customer support agent during a call with a customer. The deployed system was therefore required to be fast (less than 200ms inference time) but economical (able to operate on CPU, rather than more expensive GPUs).

 To address the above issues, in this paper, we present a simple yet effective method, \textit{distill-then-fine-tune}, to transfer knowledge from a large and complex model to a small and simple model while reaching a similar performance as the large model. More specifically, we fine-tune a state-of-the-art NER model, LUKE~\cite{DBLP:conf/emnlp/YamadaASTM20}, on our limited amount of noisy telephone conversations and predict the labels of a large amount of unlabeled conversations, denoted as distillation data. The smaller model is then trained on the distillation data using pseudo-labels.  We conduct extensive experiments with our proposed approach and observe that our distilled model achieves 75x inference speed boost while reserving $99.09\%$ F1 score of its teacher. This makes our proposed approach very effective in limited budget scenarios as it does not require the annotation of a huge amount of noisy data that would otherwise be required to fine-tune simpler
 transformers on downstream tasks.

% {\color{blue} motivation (human annotation costly) and contributions}
%\vspace{-2mm}
\section{Related Work}
%\vspace{-2mm}

% \subsection{Knowledge Distillation}

% Knowledge distillation~\cite{DBLP:journals/corr/HintonVD15}  can be viewed as a way to compress a large model into a small model. In knowledge distillation, the student (small model) is trained to mimic the behavior of the teacher (large model). Typically, the complete training procedure has two steps. The teacher model is first trained on a set of annotated training examples. And the teacher model is used to generate predicted distribution of training labels for the student model to learn. 

% In knowledge distillation, the student does not have to have the same architecture as the teacher. There is no dependency between the teacher and the student. The main benefit of this is that one can select a very simple model despite how complex the teacher model is. This also makes this approach easy to implement. 

% \cite{DBLP:journals/corr/abs-1910-01108} proposed a method to pre-train a smaller general purpose language representation model, which can then be fine-tuned with good performances on a wide range of downstream tasks. Raphael et al, 2019 proposed a task specific knowledge distillation method to show that lightweight neural networks can still be made competitive without additional annotated data, or input features. More importantly, their experiments showed that using an additional unlabeled transfer dataset can augment the training set for more effective knowledge transfer. 

% \subsection{Named Entity Recognition}

NER is often framed as a sequence labeling problem~\cite{DBLP:journals/corr/HuangXY15, DBLP:conf/coling/AkbikBV18} where a model is used to predict the entity type of each token. Previously, various models based on the recurrent neural network architecture have been widely used for this task. In recent years, pre-trained language models have been employed to perform the NER task where a new prediction layer is added into the pre-trained model to fine-tune for sequence labeling ~\cite{devlin2019bert}.

More recently, ~\cite{DBLP:conf/emnlp/YamadaASTM20} proposed a new approach to provide the contextualized representations of words and entities based on a bidirectional transformer. In their proposed model, LUKE, they treat words and entities in a given context as independent tokens, and output the contextualized representations of them. The LUKE model achieved impressive performance in various entity-related tasks. However, this model is inherently slow due to its complex  architecture and so it is not applicable for usage in production environments in a limited computational budget scenario.

In scenarios where the computational budget is limited, using a smaller model that can mimic the behaviour of the large model can be used. Knowledge distillation~\cite{DBLP:journals/corr/HintonVD15} is one such technique where a large model is compressed into a small model. % Significant work has been done on knowledge distillation in recent years where transformer-based architecture has been used due to its impressive performance \cite{gou2021knowledge}. 
One prominent approach for Knowledge Distillation that has been used in recent years is the work of ~\cite{DBLP:journals/corr/abs-1903-12136}, where they proposed a task specific knowledge distillation method to show % that lightweight neural networks can still be made competitive without additional annotated data, or input features. More importantly, their experiments showed 
that using an additional unlabeled transfer dataset can augment the training set for more effective knowledge transfer.  However, most prior work that leveraged such knowledge distillation techniques focused on typed input, whereas the amount of work that leveraged knowledge distillation for noisy texts (e.g., telephone conversation transcripts) is very limited \cite{gou2021knowledge}. Motivated by the advantages of knowledge distillation, in this work, we also leverage  knowledge distillation to address the computational issues that occur while utilizing large state-of-the-art language models in a limited computational environment, while minimizing the amount of noisy data that must be human-annotated for use during fine-tuning. 

%\vspace{-2mm}

\section{Datasets}
%\vspace{-2mm}
In this section, we first introduce the in-domain training data (noisy human-to-human conversations) that we sampled and annotated to train the teacher model. Then, we describe the data used for knowledge distillation of the student model. 

\begin{table}[t!]
\centering
\small
\begin{tabular}{|c|c|c|c|c|}
\hline
\textbf{Type} & \textbf{Utterances} & \textbf{Person} & \textbf{Prod/Org} & \textbf{Location} \\ \hline
Train              & 16124               & 4852            & 4443              & 4135         \\ \hline
Dev                & 2292                & 682             & 627               & 629          \\ \hline
Test               & 4497                & 1382            & 1274              & 1151         \\ \hline
\end{tabular}
\caption{Labeled in-domain dataset class distribution. The numbers under each entity type represent number of utterances containing the specific type.}
\label{tab:data_distribution}
%\vspace{-2mm}
\end{table}

% Train              & 16124               & 6691            & 5493              & 4136         \\ \hline
% Dev                & 2292                & 682             & 627               & 629          \\ \hline
% Test               & 4497                & 1382            & 1274              & 1151         \\ \hline

%\vspace{-4mm}

\subsection{In-domain Data Annotation}
%\vspace{-2mm}
Since our in-domain dataset is sampled from transcripts produced by an ASR system, the dataset does not contain any punctuation marks and only contains partial casing information. This makes the property of our dataset fundamentally different from the data that most pre-trained models are trained on. This also makes the task more difficult since upper-cased words are a very strong hint of a token being a named entity~\cite{DBLP:conf/emnlp/MayhewTR19}. 

For data annotation, we sampled 26,000 utterances from telephone conversation transcripts and had them annotated by Appen\footnote{\url{https://appen.com/}, accessed on January 4, 2022.}. Four types of named entities were labeled by the annotators: \textit{person name}, \textit{product or organization}, \textit{geopolitical location}, and \textit{none}. The detailed statistic of this dataset labeled by Appen is shown in Table~\ref{tab:data_distribution}.

%The breakdown of the named entity class distribution is shown in Table~\ref{tab:data_distribution}.

% In this work, we mainly focused on four entity types, \textbf{O}, \textbf{PERSON}, \textbf{PRODUCT}, \textbf{ORGANIZATION} and \textbf{GPE}. Of these, \textbf{O} means not a named entity and \textbf{GPE} means geopolitical entities. Further, we combined \textbf{PRODUCT} and \textbf{ORGANIZATION} to become \textbf{PRODORG} because the difference is often subtle in our domain, which we plan to explore in the future. Our in-house applied scientists and linguists labeled a set of examples for use as test questions for Appen annotators. Before entering as well as working on the job, Appen annotators need to maintain $70\%$ accuracy on the test questions. We collected final results from Appen, and Table~\ref{tab:data_distribution} shows the statistics of each interested named entities types.

 % The model now has to distinguish the tokens inside a named entity from those are outside from contexts along.  

% Here, `\#' denotes `total number of'.

\begin{figure*}[t!]
\begin{center}
  \includegraphics[width=1\linewidth]{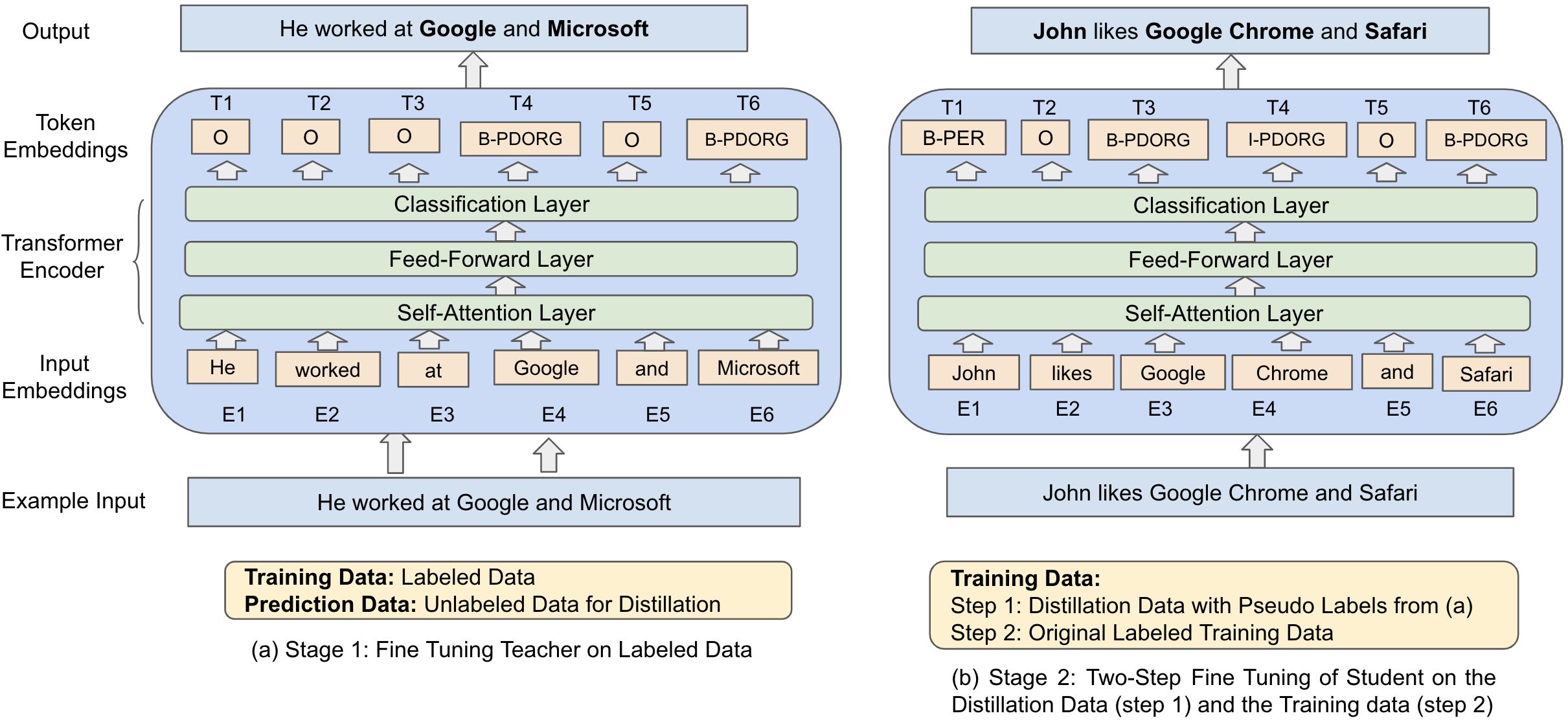}
  \caption{Our knowledge distillation approach: (a) first, fine-tune the teacher model (LUKE) on the labeled dataset, and generate the pseudo-labels of a huge amount of unlabeled data for distillation. (b) Next, fine-tune the student model (DistilBERT) in two steps, \textbf{step 1:} on the distillation data having pseudo-labels that were generated in the previous step, and \textbf{step 2:} on the original labeled training data where the teacher model was also trained. Here, `PDORG' denotes `PROD/ORG', while `Bold' font in the output layer denotes the entities tagged by the model.}
  \label{fig:model}
 \end{center}
\end{figure*} 
%\vspace{-2mm}
\subsection{In-domain Distillation Data}
%\vspace{-2mm}
Our goal is to reduce the amount of human annotated data in the training set. For this purpose, we perform knowledge distillation that transfers knowledge from a large and complex teacher model to a small and simple student model. Since the student model is expected to be much simpler than the teacher model, it requires a large amount of labeled training data. 
%We therefore obtained unlabeled utterances from business telephone conversation transcripts. 
In addition, due to the sparsity of named entities, the model cannot learn too much from randomly sampled utterances where most of them may not contain any named entities. We address this issue by using the spaCy\footnote{\url{https://spacy.io/api/entityrecognizer}} NER model to select utterances that are highly likely to contain at least one named entity of a type we are interested in. Specifically, we only used four entity types relevant to this study from the spaCy model: PERSON, ORG, GPE, PRODUCT. This sampling method produced $483,766$ unlabeled utterances from business telephone conversation transcripts and largely increased the information density in the data. However, annotating this huge amount of unlabeled data would be a prohibitively costly process. To tackle this problem, we use the trained teacher model to predict the labels of these utterances. In this way, the teacher model provides the pseudo-labels of a large unlabeled noisy dataset to alleviate the need of human annotation for such data. We use this large noisy speech data with pseudo-labels as the distillation data to train the student model. The statistics of this dataset is listed in Table~\ref{tab:pseudo_data_distribution}.

\begin{table}[t!]
\centering
\small
\begin{tabular}{|c|c|}
\hline

\textbf{Type} & \textbf{\# Examples} \\ \hline
\textit{Positive} {utterances} & 347,412  \\ \hline
\textit{Negative} {utterances} & 136,354 \\ \hline
{Utterances containing} \textit{Person} {tags}   & 179,495 \\ \hline
{Utterances containing} \textit{Prod/Org} {tags}  & 97,857   \\ \hline
{Utterances containing} \textit{Location} {tags} &  138,989 \\ \hline

\end{tabular}
\caption{Pseudo-labeled distillation data class distribution. ``Positive utterances" are those that contain any of the 3 entity types, and ``Negative utterances" are those that do not contain any of the 3 entity types. Here, `\#' denotes `Total number of'. }
\label{tab:pseudo_data_distribution}
%\vspace{-2mm}
\end{table}

 %\vspace{-2mm}

\section{Our Proposed Approach} 
%\vspace{-2mm}
In this section, we first describe the architectures of the teacher and student models. We then describe our proposed knowledge distillation method, \textit{distill-then-fine-tune}, that can be broken down into four steps: i) fine-tune the teacher model on the in-domain data, ii) sample distillation data from % a pool of 
unlabeled examples, iii) perform distillation, and iv) fine-tune the student model. An overview of our proposed approach is illustrated in Figure~\ref{fig:model}. 

%\vspace{-3mm}
\paragraph{Model Architecture:}

We use LUKE, a bidirectional transformer, that was pre-trained by \cite{DBLP:conf/emnlp/YamadaASTM20} on Wikipedia data to learn contextualized representations of words and entities. In LUKE, the input representation of a token (word or entity) is computed using three types of embedding: token embedding, position embedding, and entity type embedding. Token embedding, which is decomposed into two small matrices, represents the corresponding token. Position embedding represents the position of a token in a word sequence, while the entity type embedding represents whether the token is an entity. 
To further leverage the entity type embedding, an entity-aware self attention mechanism is used to handle interactions between entities in a given word sequence. Since LUKE is a large model that contains approximately 483M parameters (355M on its encoder and 128M for entity embeddings), we use it as the teacher to teach a student model. % Note that the entity vocabulary of LUKE is also huge, as it contains about 500K entities. 

For the student model, we adapt the DistilBERT~\cite{DBLP:journals/corr/abs-1910-01108} model, a 6-layer bidirectional transformer encoder that was pre-trained for the language modeling task by \citet{DBLP:journals/corr/abs-1910-01108}. The DistilBERT model was initialized from its teacher BERT model by taking one layer out of two. It was pre-trained on the same corpus as BERT while using both the distillation loss and the masked language modelling loss. It contains approximately 66M parameters (approximately one seventh the size of the teacher model), making it more economical to deployment in a production environment with limited resources.
%\vspace{-2mm}
\paragraph{Distillation Method:}

Our goal is to build an NER system that can detect named entities in business conversations, but the LUKE model that we employ as a teacher model was pre-trained on written text, which is very different from noisy transcribed human-to-human conversations. To adapt to the domain of business conversations, we first fine-tune the LUKE model on 16,124 in-domain human-annotated examples (see Section 3.1 for details). The resulting model is called LUKE\textsubscript{ft}. The LUKE\textsubscript{ft} model serves as the teacher that generates pseudo-labels for the distillation data (see Section 3.2 for details).

Next, we use a two-step fine-tuning approach for the student model \cite{fu2021improving,laskar2022domain}. The student model is initialized with the pre-trained DistilBERT model. For step 1, we fine-tune the student model on the distillation data with pseudo-labels generated by the teacher. During the training stage, we use the cross entropy loss defined below. % in Equation~\ref{eqn:cross_entropy}:
%\vspace{-2mm}
\begin{equation}
\label{eqn:cross_entropy}
L_{CE} = - \frac{1}{N} \sum_{n=1}^{N}\log\frac{e^{\hat{y}_{n, y_n}}}{\sum_{c=1}^{C}e^{\hat{y}_{n,c}}}
\end{equation}
Here, $N$ is the number of samples in a batch, and $C$ denotes the number of classes. $\hat{y}_{n, c}$ is the logit of the $c$-th class in the $n$-th example, and $\hat{y}_{n, y_n}$ is the logit of the gold class in the $n$-th example.
%%\vspace{-2mm}

For the final distillation step, we fine-tune the student model further on the in-domain human-annotated data. The resulting child model is termed \textbf{DistilBERT\textsubscript{dtft}}.

% \section{Experimental Settings}

%\subsection{Implementation Details}
%[{\color{blue}{WE MAY MENTION THE IMPLEMENTATION DETAILS SIMILAR TO THIS WORK (SEE SECTION 4.4: https://aclanthology.org/2020.lrec-1.676.pdf)}}]
%I will mention the implementation library in model deployment. 
%}

% \textbf{Number of layers:} To increase the inference speed, we experiment with reducing the number of Transformer layers of the model. We train 6-layer and 2-layer models as students for comparison. 

% \textbf{In-domain fine-tuning:} We also want to answer the question “If fine-tuning can further improve the performance of a distilled model”. To do that, we compare models that are fine-tuned on our in-domain data with models that are not. 

\begin{table}[t!]
\small
\centering
\begin{tabular}{|c|c|c|}
\hline
\textbf{Model}          & \textbf{F1 Score} & \textbf{Inference Time}  \\ \hline
LUKE\textsubscript{ft}             & 86.07            & 2980ms             \\ \hline
DistilBERT\textsubscript{ft}       & 83.08            & 40ms                    \\ \hline
\textbf{DistilBERT\textsubscript{dtft}}  & 85.29            & 40ms                            \\ \hline
% \textbf{DistilBERT\textsubscript{dtft}} - 2-layer  & 85.33            & 14ms                            \\ \hline

\end{tabular}
\caption{Performance of our proposed DistilBERT\textsubscript{dtft} models (fine-tuned on a large amount of distillation data and a small amount of in-domain human-annotated data) compared to the LUKE\textsubscript{ft} and  DistilBERT\textsubscript{ft} models that were fine-tuned only on the in-domain human-annotated data. Inference time is measured on a 2.20Ghz Intel Xeon CPU with sixteen virtual cores.}
\label{tab:benchmarking}
\end{table}

%\begin{table*}[t!]
%\centering
%\begin{tabular}{|c|c|c|}
%\hline
%\textbf{Model} & \textbf{F1 Score} & \textbf{Inference time}  \\ \hline
%LUKE (Baseline) &    86.07   &   2980 ms                \\ \hline
%DistilBERT (Baseline)   &  83.08  & 40 ms                  \\ \hline
%6 layer DistilBERT\textsubscript{dtft} &  85.29  & 40 ms                     \\ \hline
%2 layer DistilBERT\textsubscript{dtft} & 85.33     & 14 ms                \\ \hline
%\end{tabular}
%\caption{Ablation test result based on layer reduction.}
%\label{tab:inference_time}
%\end{table*}

%\begin{table*}[t!]
%\centering
%\begin{tabular}{|c|c|c|c|}
%\hline
%\textbf{Model} & \textbf{F1 Score} & \textbf{FT on Distillation Data} & \textbf{FT on Labeled Training Data}  \\ \hline
%LUKE (Baseline) &    86.07   &  NO & YES                \\ \hline
%DistilBERT (Baseline)   &  83.08  & NO & YES                  \\ \hline

%6 layer DistilBERT &  84.59  & YES & NO                   \\ \hline

%2 layer DistilBERT & 84.56    & YES & NO                 \\ \hline
%6 layer DistilBERT\textsubscript{dtft} &  85.29  & YES & YES                    \\ \hline
%2 layer DistilBERT\textsubscript{dtft} & 85.33  & YES & YES                \\ \hline

%\end{tabular}
%\caption{Ablation test result based on fine-tuning (FT) steps.}
%\label{tab:finetuning_steps}
%\end{table*}
%\vspace{-1mm}
\section{Experiments}
%\vspace{-1mm}

In this section, we describe our experimental settings and results. 

% In addition, we evaluate a 2-layer variant of the DistilBERT\textsubscript{dtft} model.

%\vspace{-1mm}
\subsection {Experimental Settings}
%\vspace{-2mm}
Below, we discuss the baseline models and the training parameters used in our experiments.

\textbf{Baselines:} To compare the performance with our proposed model, we use the following baselines, 
\textbf{\textit{(i) LUKE\textsubscript{ft}}}: The pre-trained LUKE model fine-tuned on our human-annotated in-domain training data, and \textbf{\textit{(ii) DistilBERT\textsubscript{ft}}}: Similar to the other baseline, it was fine-tuned only on our human-annotated in-domain training data. 

\textbf{Training Parameters:} For the teacher model, LUKE\textsubscript{ft}, we set the batch size to 2, learning rate to $5\times10^{-5}$, and the number of epochs to 3. For the student DistilBERT model, we set the batch size to 32 and the learning rate to $5\times10^{-5}$, and the number of epochs to 5. 

% \nop{ Below, we compare its performance with the baseline models.

% \subsection{Effectiveness of the Proposed Model}

% We show the results of our proposed DistilBERT\textsubscript{dtft} model as well as the baselines in Table \ref{tab:benchmarking}.
%\vspace{-1mm}
\subsection {Results and Analyses}
%\vspace{-2mm}
% \textbf{Effectiveness in terms of the F1 score:} 

From Table \ref{tab:benchmarking}, we see that the LUKE\textsubscript{ft} model (fine-tuned on in-domain human-annotated data) achieves the highest F1 score, 86.07\%, but with an inference time of 2980ms it is not practical for realtime applications.

The DistilBERT\textsubscript{ft} model (also fine-tuned only on the in-domain human-annotated data), with an inference time of 40ms is suitable for realtime application, but loses almost three percentage points of accuracy, reducing to an F1 score of 83.08\%.

Our proposed \textbf{DistilBERT\textsubscript{dtft}} model, which leverages two stage of fine-tuning (uses the large distillation data on stage 1 of fine-tuning and the human-annotated data on stage 2 of fine-tuning) brings the F1 score back to within $1\%$ of the LUKE\textsubscript{ft} model. Since \textbf{DistilBERT\textsubscript{dtft}} model has the same model architecture and the same number of parameters as the DistilBERT\textsubscript{ft} model, its inference time is identical: 40ms, i.e. 75x faster than LUKE\textsubscript{ft}. This makes \textbf{DistilBERT\textsubscript{dtft}} model applicable for production deployment as it achieves an improved F1 score with high efficiency while requiring less computational resources due to its small size. 

\section{Conclusion}
%\vspace{-2mm}
In this paper, we introduce the \textit{distill-then-fine-tune} method for entity recognition on real world noisy data to deploy our NER model in a limited budget production environment. By generating pseudo-labels using a large teacher model pre-trained on typed text while fine-tuned on noisy speech text to train a smaller student model, we make the student model 75x times faster while reserving 99.09\% of its accuracy. These findings demonstrate that our proposed approach is very effective in limited budget scenarios to alleviate the need of human labeling of a large amount of noisy data. % The student model outperforms a model that has the same number of parameters by $2.21\%$ absolute F1 score. % We also observe that the size of the student model could be further reduced without any loss in accuracy while the size of distillation data is big enough. 
In the future, we will explore how to apply knowledge distillation to other tasks \cite{laskar2022auto, laskar-etal-2022-blink, khasanova-etal-2022-developing} containing noisy data.

\section*{Ethics Statement}
% Scientific work published at EMNLP 2022 must comply with the \href{https://www.aclweb.org/portal/content/acl-code-ethics}{ACL Ethics Policy}. We encourage all authors to include an explicit ethics statement on the broader impact of the work, or other ethical considerations after the conclusion but before the references. The ethics statement will not count toward the page limit (8 pages for long, 4 pages for short papers).
The data used in this research is comprised of individual sentences that do not contain sensitive, personal, or identifying information. Each machine-sampled utterance is labelled by annotators before the utterance is used as part of the training dataset.  While annotator demographics are unknown and therefore may introduce potential bias in the labelled dataset, the annotators are required to pass a screening test before completing any labels used in these experiments, thereby mitigating this unknown to some extent. Future work should nonetheless strive to improve training data further in this regard.

%\section{Language Resource References}
%\label{lr:ref}
% \bibliographystylelanguageresource{lrec2022-bib}
% \bibliographylanguageresource{languageresource}

\bibliography{anthology,custom}
\bibliographystyle{acl_natbib}

\end{document}